\documentclass{article} 
\usepackage{iclr2023_conference,times}

\usepackage[utf8]{inputenc} 
\usepackage[T1]{fontenc}    
\usepackage{hyperref}       
\usepackage{url}            
\usepackage{booktabs}       
\usepackage{amsfonts}       
\usepackage{nicefrac}       
\usepackage{microtype}      
\usepackage{xcolor}         
\usepackage{float}

\newfloat{figtab}{htb}{fgtb}

\usepackage{times}
\usepackage{soul}
\usepackage{url}
\usepackage[small]{caption}
\usepackage{graphicx}
\usepackage{amsmath}
\usepackage{booktabs}
\usepackage{algorithm}
\usepackage{algorithmic}
\usepackage{enumitem}
\usepackage{amsmath}
\usepackage{amssymb}
\usepackage{mathtools}
\usepackage{amsthm}
\usepackage{times}
\usepackage{soul}
\usepackage{url}
\usepackage{times}
\usepackage{soul}
\usepackage{url}
\usepackage{bm}
\usepackage{subfigure}

\newcommand{\plcr}[0]{\textsc{Plcr}}
\newcommand{\pico}[0]{\textsc{Pico}}
\newcommand{\cavl}[0]{\textsc{Cavl}}
\newcommand{\lws}[0]{\textsc{Lws}}
\newcommand{\vpll}[0]{\textsc{Valen}}
\newcommand{\rcpll}[0]{\textsc{RC}}
\newcommand{\ccpll}[0]{\textsc{CC}}
\newcommand{\proden}[0]{\textsc{Proden}}

\newcommand{\plmap}[0]{\textsc{Idgp}}
\newcommand{\plml}[0]{\textsc{Idgp-ml}}

\newtheorem{theorem}{Theorem}
\newtheorem{lemma}{Lemma}
\newtheorem{definition}{Definition}

\title{Decompositional Generation Process for Instance-Dependent Partial Label Learning}

\author{%
	Congyu~Qiao, Ning~Xu$^{*}$, Xin~Geng\thanks{Corresponding authors.} \\
	School of Computer Science and Engineering, Southeast University, Nanjing 210096, China\\
	\texttt{\{qiaocy, xning, xgeng\}@seu.edu.cn} 
}

\iclrfinalcopy 
\begin{document}

\maketitle

\begin{abstract}
Partial label learning (PLL) is a typical weakly supervised learning problem, where each training example is associated with a set of candidate labels among which only one is true. Most existing PLL approaches assume that the incorrect labels in each training example are randomly picked as the candidate labels and model the generation process of the candidate labels in a simple way.  However, these approaches usually do not perform as well as expected due to the fact that the generation process of the candidate labels is always instance-dependent. Therefore, it deserves to be modeled in a refined way.  In this paper, we consider instance-dependent PLL and assume that the generation process of the candidate labels could decompose into two sequential parts, where the correct label emerges first in the mind of the annotator but then the incorrect labels related to the feature are also selected with the correct label as candidate labels due to uncertainty of labeling. Motivated by this consideration, we propose a novel PLL method that performs Maximum A Posterior (MAP) based on an explicitly modeled generation process of candidate labels via decomposed probability distribution models. Extensive experiments on manually corrupted benchmark datasets and real-world datasets validate the effectiveness of the proposed method. Source code is available at \url{https://github.com/palm-ml/idgp}.
\end{abstract}

\section{Introduction}

Partial label learning (PLL) aims to deal with the problem where each instance is provided with a set of candidate labels, only one of which is the correct label.  The problem of learning from partial label examples naturally arises in a number of real-world scenarios such as web data mining \cite{luo2010learning}, multimedia content analysis \cite{zeng2013learning,chen2017learning}, and ecoinformatics \cite{liu2012conditional,tang2017confidence}.

A number of methods have been proposed to improve the practical performance of PLL. Identification-based PLL approaches \cite{jin2002learning,nguyen2008classification,liu2012conditional,chen2014ambiguously,yu2016maximum} regard the correct label as a latent variable and try to identify it. Average-based approaches \cite{hullermeier2006learning,cour2011learning,zhang2015solving} treat all the candidate labels equally and average the modeling outputs as the prediction. In addition, risk-consistent methods \cite{feng2020provably, wen2021leveraged} and classifier-consistent methods \cite{lv2020progressive,feng2020provably} are proposed for deep models. Furthermore, aimed at deep models, \cite{wang2022pico} investigated contrastive representation learning, \cite{zhang2021exploiting} adapted the class activation map and \cite{wu2022revisiting} revisited consistency regularization in PLL.

It is challenging to avoid overfitting on candidate labels, especially when the candidate labels depend on instances. Therefore, the previous methods assume that the candidate labels are instance-independent. Unfortunately, this often tends to be the case that the incorrect labels related to the feature are more likely to be picked as candidate label set for each instance. Recent work \cite{xu2021instance} has also shown that the presence of instance-dependent PLL imposes additional challenges but is more realistic in practice than the instance-independent case.

In this paper, we focus on the instance-dependent PLL via considering the essential generating process of candidate labels in PLL. To begin with, let us rethink meticulously how candidate labels arise in most manual annotation scenarios. When one annotates an instance, though the correct label has already emerged in the mind of the annotator first, the incorrect labels which are related to the feature of the instance confuse the annotator, then leading to the result that the correct label and some incorrect labels are packed together as the candidate labels. Therefore, the generating process of the candidate labels in instance-dependent PLL could be decomposed into two stages, i.e., the generation of the correct label of the instance and the generation of the incorrect labels related to the instance, which could be described by Categorical distribution and Bernoulli distribution, respectively.

Motivated by the above consideration, we propose a novel PLL method named {\plmap}, i.e., Instance-dependent partial label learning via Decompositional Generation Process.  Before performing {\plmap}, the distribution of the correct label and the incorrect label in the candidate label set of each training example should be modeled explicitly by decoupled probability distributions Categorical distribution and Bernoulli distribution. Then we perform Maximum A Posterior (MAP) estimation on the PLL training dataset to deduce a risk minimizer. To optimize the risk minimizer, Dirichlet distribution and Beta distribution are leveraged to model the condition prior inside and estimate the parameters of Categorical distribution and Bernoulli distribution due to the conjugacy. Finally, we refine prior information by updating the parameters of the corresponding conjugate distributions iteratively to improve the performance of the predictive model in each epoch. Our contributions can be summarized as follows:
\begin{itemize}[topsep=0pt,leftmargin=10pt,itemsep=0pt]
	\item We for the first time explicitly model the generation process of candidate labels in instance-dependent PLL. The entire generating process is decomposed into the generation of the correct label of the instance and the generation of the incorrect labels, which could be described by Categorical distribution and Bernoulli distribution, respectively. 
	\item We optimize the models of Categorical distribution and Bernoulli distribution via the MAP technique, where the corresponding conjugate distributions, i.e., Dirichlet distribution and Beta distribution are induced.
	\item We derive an estimation error bound of our approach, which demonstrates that the empirical risk minimizer would approximately converge to the optimal risk minimizer as the number of training data grows to infinity.  
\end{itemize}
Experiments on benchmark and real-world datasets validate the effectiveness of our proposed method.

\section{Related Work}

In this section, we briefly review the literature for PLL from two aspects, i.e., traditional PLL and deep PLL. The former absorbs many classical machine learning techniques and usually utilizes linear models while the latter embraces deep learning and builds upon deep neural networks. We focus on the underlying assumptions on the generation of candidate labels behind part of them. 

Traditional PLL usually uses average-based or identification-based approaches to disambiguate the candidate label set. Average-based approaches treat each candidate label of the instance equally \cite{hullermeier2006learning,cour2011learning,zhang2015solving}. Typically, \cite{hullermeier2006learning,zhang2015solving} apply the K-nearest neighbor technique to predict a new instance through voting. Identification-based approaches are constantly trying to identify the possible correct label, either explicitly or implicitly during the training phase, in order to reduce label ambiguity \cite{jin2002learning,nguyen2008classification,liu2012conditional,chen2014ambiguously,yu2016maximum}. For instance, \cite{nguyen2008classification,yu2016maximum} formulate their objective functions by treating the correct label as a latent variable via the maximum margin criterion. \cite{zhang2016partial,feng2018leveraging,wang2019adaptive,xu2019partial} iteratively update the confidence of each candidate label or the label distribution via leveraging the topological information in the feature space. Note that most traditional PLL methods do not take the generation process of candidate labels into consideration. \cite{liu2012conditional} propose the Logistic Stick-Breaking Conditional Multinomial Model to depict the generation process but simply assume the candidate labels to be instance-indpendent.

Deep PLL has been studied recently and advanced the practical application of PLL, where the PLL approaches are not limited to linear models and low-efficiency optimization. \cite{yao2020deep} pioneer the use of deep convolutional neural networks and employ a regularization term of uncertainty and a temporal-ensembling term to train the deep model. \cite{lv2020progressive} propose a progressive identification method that allows PLL to be compatible with arbitrary models and optimizers and  for the first time, performs impressively on image classification benchmarks. \cite{yao2020network} introduce a network-cooperation mechanism in deep PLL which trains two networks simultaneously to reduce their respective disambiguation errors. \cite{feng2020provably} derive a risk-consistent estimator via employing the importance reweighting strategy and a classifier-consistent estimator via utilizing the transition matrix for PLL. \cite{wen2021leveraged} introduce a leverage parameter that weights losses on partial labels and non-partial labels. \cite{xu2021instance} apply variational inference to iteratively recover label distribution for each instance. \cite{lyu2022deep} transform the PLL problem into an “instance-label” matching selection problem. \cite{he2022partial} utilize semantic label representations to enhance the label disambiguation. \cite{wang2022pico} build a contrastive learning framework for PLL to perform disambiguation. \cite{zhang2021exploiting} discover class activation map could be utilized for disambiguation and further purpose class activation value to capture the learned representation information in a more general way. \cite{wu2022revisiting} employ manifold consistency regularization to preserve the manifold structure both in feature and label space. 

From \cite{feng2020provably}, the generation process of candidate labels has been paid attention to in Deep PLL. \cite{feng2020provably} make a naive assumption that the candidate label set is uniformly sampled for each instance. \cite{wen2021leveraged} extend the uniform case to the class-dependent one but remain instance-independent. \cite{xu2021instance} for the first time consider PLL under the instance-dependent case, a more realistic but challenging setting. However, \cite{xu2021instance} do not explicitly model the generation process of instance-dependent candidate labels. The generative model in it is employed only to generate label distribution. In this paper, we not only consider the instance-dependent PLL but also explicitly model the generation process of instance-dependent candidate labels via decoupled probability distributions. 

\section{Proposed Method}
 
First of all, we introduce the necessary notations for our approach. In PLL, the labeling information of every instance in the training dataset is corrupted from a correct label into a candidate label set, which contains the correct label and incorrect candidate labels but is not a complete label set. Let $\mathcal{X} \subseteq \mathbb{R}^{q}$ be the $q$-dimensional feature space of the instances and $\mathcal{Y} = \{1,2,\dots,c\}$ be the label space with $c$ class labels. Then a PLL training data set can be formulated as $\mathcal{D} = \{(\bm{x}_{i}, S_{i})| 1 \leq i \leq n\}$, where $\bm{x}_i\in \mathcal{X}$ denotes the $i$-th $q$-dimensional feature vector and $S_i\in \mathcal{C}=2^{\mathcal{Y}} \setminus \{ \emptyset, \mathcal{Y} \}$ is the candidate label set annotated for $\bm{x}_i$. $S_i$ consists of the correct label $y_i\in\mathcal{Y}$ and the incorrect candidate label set $\overline{S}_i^{y_i} = S_i\setminus\{y_i\}$. We aim to find a multi-class classifier $f: \mathcal{X} \mapsto \mathcal{Y}$ according to $\mathcal{D}$. 

\subsection{Overview}

To deal with the instance-dependent PLL problem, we introduce the explicit generation model of instance-dependent candidate labels, which decouples the distribution of the correct and incorrect candidate labels. Then Category distribution and Bernoulli distribution are leveraged to depict them respectively with Dirichlet distribution and Beta distribution as their prior distributions. Based on the probabilistic generation model, we simply deduce the log-likelihood loss function for optimization and then move forward to the MAP optimization problem in consideration of the prior distributions Dirichlet and Beta.  Finally, we further propose the algorithm {\plmap}, which keeps and refines the prior information.

\subsection{Generation Model of Candidate Labels}

Focusing on instance-dependent partial label learning, we propose a novel generation model of candidate labels, which explicitly models the generation process and decouples the distribution of the candidate labels into the distribution of the correct label and incorrect candidate labels respectively. To form a candidate label set, we suppose the correct label is first selected according to a posterior distribution dependent on the instance. And then the incorrect candidate labels related to the feature of instance emerge to disturb the annotator and are sampled from another posterior distribution dependent on the instance. The concrete generation model is demonstrated as follows.

Given an instance $\bm{x}_i$, its candidate label set $S_i$, consisting of the correct label $y_i$ and the incorrect candidate labels $\overline{S}_i^{y_i}$, is drawn from a probability distribution with the following density:
\begin{equation}\label{generation_model}
	\begin{aligned}
		p(S_i|\bm{x}_i) = \sum_{j\in S_i}p(y_i=j|\bm{x}_i)p(\overline{S}^j_i|\bm{x}_i).
	\end{aligned}
\end{equation}
Here, $p(S_i|\bm{x}_i)$ suggests our generation model is entirely dependent on the instance. Our generation model builds upon the PLL assumption that the correct label is always in the candidate set, which allows for directly splitting $S_i$ into $y_i$ and $\overline{S}_i^{y_i}$. For the given candidate label set $S_i$, each label $j\in S_i$ have the possibility $p(y_i=j|\bm{x}_i)$ sampled by the annotator as the correct label in the first stage of generation and the incorrect candidate labels $\overline{S}^j_i$ are then sampled with the possibility $p(\overline{S}^j_i|\bm{x}_i)$ in the second stage to form the entire candidate label set $S_i$. The generation of the correct label and incorrect labels are conditioned on the instance $\bm{x}_i$. In this way, we decompose the generation process of the instance-dependent candidate labels. 

After the decomposition of generation, the corresponding probability distributions are leveraged to depict $p(y_i|\bm{x}_i)$ and $p(\overline{S}^{y_i}_i|\bm{x}_i)$. We assume that the correct label $y_i$ of each instance $\bm{x}_i$ is drawn from a Categorical distribution with the parameters $\bm{\theta}_i$, where $\bm{\theta}_i=[\theta_i^1, \theta_i^2, \dots, \theta_i^c]\in[0, 1]^c$ is a $c$-dimension vector with the constraint $\sum_{j=1}^{c}\theta_i^j=1$, i.e., 
\begin{equation}\label{c_d}
	p(\bm{l}_i|\bm{x}_i, \bm{\theta}_i)=Cat(\bm{l}_i|\bm{\theta}_i)=\prod_{j=1}^{c}(\theta_i^j)^{l^j_i},
\end{equation}
where $\bm{l}_i=[l_i^{1}, l_i^{2}\dots, l_i^{c}]\in\{0,1\}^c$ to represents whether the $j$-th label is the correct label $y_i$, i.e.,$l_i^{j} = 1$ if $j=y_i$, otherwise $l_i^j=0$. 

Besides, for the incorrect candidate label set $\overline{S}_i^{y_i}$, it can also be denoted by a logical label vector $\overline{\bm{s}}_i^{y_i} = [\overline{s}_i^1, \overline{s}_i^2\dots, \overline{s}_i^c]\in\{0, 1\}^c$ to represent whether the $j$-th label is the incorrect candidate label, i.e., $\overline{s}_i^j = 1$ if $ j\in \overline{S}_i^{y_i}$, $\overline{s}_i^j=0$ if $ j\notin \overline{S}_i^{y_i}$. In order to describe $p(\overline{S}^{y_i}_i|\bm{x}_i)$ with a probabilistic model, we further decouple the incorrect candidate labels by assuming that the distribution of each variable $\overline{s}^j_i$ is independent from each other, i.e, $p(\overline{\bm{s}}^{y_i}_i|\bm{x}_i) = \prod_{j\in \overline{S}^{y_i}_i}p(\overline{s}^j_i|\bm{x}_i)\prod_{j\notin \overline{S}^{y_i}_i}(1 -p(\overline{s}^j_i|\bm{x}_i))$ and the incorrect candidate label set $\overline{S}^{y_i}_i$ is drawn from a multivariate Bernoulli distribution with the parameters $\bm{z}_i$, where $\bm{z}_i=[z_i^1,z_i^2,\dots, z_i^c]\in[0, 1]^c$ is a c-dimension vector, i.e., 
\begin{equation}\label{b_d}
	p(\overline{\bm{s}}^{y_i}_i|\bm{x}_i, \bm{z}_i) = \prod_{j=1}^{c} Ber(\overline{s}_i^j|z_i^j)=\prod_{j=1}^{c} (z_i^j)^{\overline{s}_i^j}(1-z_i^j)^{1-\overline{s}_i^j}.
\end{equation}

For the latter estimation of $\bm{\theta}_i$ in MAP, we introduce Dirichlet distribution, the conjugate distribution of the Categorical distribution, as its conditional prior, i.e., $p(\bm{\theta}_i | \bm{x}_i)=Dir(\bm{\theta}_i|\bm{\lambda}_i)$,
where $\bm{\lambda}_i = [\lambda_i^1, \lambda_i^2, \dots, \lambda_i^c]^\top (\lambda_i^j > 0)$ is a $c$-dimensional vector as the output of our main branch inference model $f$ parameterized by $\bm{\Theta}$ for an instance $\bm{x}_i$, i.e, $\bm{\lambda}_i = a \cdot \exp\left(f(\bm{x}_i; \bm{\Theta}) / \gamma\right) + b$, where $a$, $b$ and $\gamma$($a\geq 1,b \geq 0,\gamma>0$) are used to resolve the scale ambiguity. Note that the main branch inference model $f$ is leveraged as the final predictive model to accomplish the PLL target $\mathcal{X} \mapsto \mathcal{Y}$  by using $\widehat{y}_i=\arg\max_{j}\widehat{\theta}_i^j$, where $\widehat{\theta}_i^j$ will be estimated by $\bm{\lambda}_i$ with the conjugacy later.

Likewise, to estimate $\bm{z}_i$ latter in MAP, we introduce Beta distribution, the conjugate distribution of Bernoulli distribution, as the conditional prior parameterized by $\bm{\alpha}_i=[\alpha_i^1, \alpha_i^2, \dots, \alpha_i^c]^\top$ and $\bm{\beta}_i=[\beta_i^1, \beta_i^2, \dots, \beta_i^c]^\top (\alpha_i^j > 0, \beta_i^j > 0)$, i.e., $p(\bm{z}_i|\bm{x}_i) = \prod_{j=1}^{c}Beta(z_i^j| \alpha_i^j, \beta_i^j)$. 
We employ an auxiliary branch model $g$ parameterized by $\bm{\Omega}$ to output $\bm{\Lambda}_i^\top = [\bm{\alpha}_i^\top, \bm{\beta}_i^\top]$,i.e, 
$\bm{\Lambda}_i = a \cdot \exp\left(g(\bm{x}_i; \bm{\Omega}) / \gamma \right) + b$. The same model and constants to scale $\bm{\Lambda}_i$ are implemented to simplify our approaches. It should be noted that the predictive model $f(\bm{x}_i; \bm{\Theta})$ or auxiliary model $g(\bm{x}_i; \bm{\Omega})$ can be any deep neural network once its output satisfies the corresponding constraint.

\subsection{Optimization Using Maximum A Posterior}

Based on the generation model of candidate labels denoted by Eq.(\ref{generation_model}), Eq.(\ref{c_d}) and Eq.(\ref{b_d}), by using the technique of Maximum Likelihood (ML) estimation we can immediately induce the log-likelihood loss function for the PLL training data set to be optimized:
\begin{equation}\label{L_ml}
	\begin{aligned}
		\mathcal{L}_\text{{ML}} &= - \sum_{i=1}^{n}\log{p(S_i|\bm{x}_i, \bm{\theta}_i, \bm{z}_i)}=-\sum_{i=1}^{n}\log\sum_{j\in S_i}\theta_i^j\prod_{k\in \overline{S}_i^j}z_i^k\prod_{k\notin \overline{S}_i^j}(1-z_i^k). 
	\end{aligned}
\end{equation}
The Eq.(\ref{L_ml}) demonstrates how the distribution of the correct label interacts with that of the incorrect candidate labels after decoupling the generation. In the backpropagation of the training process, they provide weight coefficients for each other.

In order to bring in prior information for PLL, we further introduce {\plmap}, which performs MAP in the training data set. In {\plmap}, we are more concerned about maximizing the joint distribution $p(\bm{\theta}, \bm{z}|\bm{x}, S)$. This will lead to the following optimization problem $\mathcal{L}_\text{{MAP}} = \mathcal{L}_\text{{ML}} + \mathcal{L}_\text{{reg}}\,$, where
\begin{equation}\label{L_reg}
	\mathcal{L}_\text{{reg}} = - \sum_{i=1}^{n}\log{p(\bm{\theta}_i|\bm{x}_i)} + \log{p(\bm{z}_i|\bm{x}_i)}.
\end{equation}
Compared to ML, our {\plmap} framework provides a natural way of leveraging prior information via optimizing the extra condition prior illustrated by the Eq.(\ref{L_reg}) in the training process, which is significant for PLL due to the implicit supervision information. Combined with the prior distribution Dirichlet and Beta, $\mathcal{L}_\text{{reg}}$ can be analytically calculated as follows:
\begin{equation}\label{L_reg_w}
	\begin{aligned}
		&\mathcal{L}_\text{{reg}} = - \sum_{i=1}^{n}\sum_{j=1}^{c}(\lambda_i^j-1)\log{\theta_i^j} +(\alpha_i^j-1)\log{z_i^j}+(\beta_i^j-1)\log{(1-z_i^j)} .
	\end{aligned}
\end{equation}
The mathematical derivations of Eq.(\ref{L_ml}) and Eq.(\ref{L_reg_w}) are provided in Appendix \ref{D_eq4_6}. By combining the Eq.(\ref{L_ml}) and the Eq.(\ref{L_reg_w}), the MAP optimization problem $\mathcal{L}$ can be calculated as follows:
\begin{equation}\label{L_map_final}
	\begin{aligned}
		&\mathcal{L} = -\sum_{i=1}^{n}\log\sum_{j\in S_i}\theta_i^j\prod_{k\in \overline{S}_i^j}z_i^k\prod_{k\notin \overline{S}_i^j}(1-z_i^k) \\
		&- \sum_{i=1}^{n}\sum_{j=1}^{c}(\lambda_i^j-1)\log{\theta_i^j} +(\alpha_i^j-1)\log{z_i^j}+(\beta_i^j-1)\log{(1-z_i^j)} .
	\end{aligned}
\end{equation}

$\mathcal{L}$ can also accommodate the uniform case, which can be seen in Appendix \ref{D_eq7}. Then, due to the conjugacy of Dirichlet and Categorical distribution \cite{minka2000estimating}, we can estimate $\theta_i^j$ by using 
\begin{equation}\label{theta_c}
	\widehat{\theta}_i^j = \mathbb{E}\left[\theta_i^j | \bm{x}_i, \bm{\lambda}_i \right]=\frac{o_i^j + \lambda_i^j}{\sum_{k=1}^{c}\lambda_i^k+o_i^k},
\end{equation}
where $o_i^j$ denotes the number of occurrences of label $j$ for $\bm{x}_i$, i.e., $o_i^j=1$ if $j\in S_i$, otherwise $o_i^j = 0$. Similarly, we can leverage the conjugacy of Beta and Bernoulli distribution to estimate $z_i^j$,i.e,
\begin{equation}\label{z_c}
	\widehat{z}_i^j = \mathbb{E}\left[z_i^j | \bm{x}_i, \alpha_i^j, \beta_i^j \right]= \frac{o_i^j + \alpha_i^j}{\alpha_i^j+\beta_i^j + o_i^j}.\\
\end{equation}
The mathematical derivations of Eq.(\ref{theta_c}) and Eq.(\ref{z_c}) are provided in Appendix \ref{D_eq8_9}.

As it is shown in the Eq.(\ref{L_reg_w}), {\plmap} provides prior information that can be included in the parameters $\lambda, \alpha$ and $\beta$ for the predictive model $f(\bm{x}; \bm{\Theta})$ and the auxiliary model $g(\bm{x}; \bm{\Omega})$ by transforming them into the weights exerted on the corresponding label. The prior information comes from the memorization effects \cite{han2020robust} of the neural network. It makes the neural network always likely to recognize and remember the correct label in priority, leading to a kind of initial disambiguation at the beginning epochs. Hence, to keep fine prior information, we replace $\bm{\lambda}_i$ with $\widehat{\bm{\lambda}}_i$ in the following way:
\begin{equation}\label{e_lambda}
	\widehat{\lambda}_{i}^{j(t)} = \left\{
	\begin{aligned}
		m\lambda_{i}^{j(r)}+&(1-m)\lambda_{i}^{j(t)},\text{if}\,  j\in S_i\\
		&1+\epsilon, \text{otherwise},\\
	\end{aligned}
	\right.
\end{equation}
where $(\cdot)^{(t)}$ denotes the vector or scalar $(\cdot)$ at the $t$ epoch, $r(\leq t)$ denote the beginning epoch we set to reserve the fine prior information for $f(\bm{x}; \bm{\Theta})$, $m\in (0,1)$ is a positive constant used to replenish the present information $\lambda_{i}^{j(t)}$ with the prior information $\lambda_{i}^{j(r)}$ and $\epsilon$ is a minor value, which means that the weight excerted on each incorrect label is negligible. In the similar way, we replace $\bm{\alpha}_i^{(t)}$ and $\bm{\beta}_i^{(t)}$ with $\widehat{\bm{\alpha}}_i^{(t)}$ and $\widehat{\bm{\beta}}_i^{(t)}$, i.e, the vectors $\widehat{\bm{\alpha}}_i^{(t)}$ and $\widehat{\bm{\beta}}_i^{(t)}$ will be calculated by
\begin{equation}\label{e_alpha_beta}
	\left\{
	\begin{aligned}
		&\widehat{\alpha}_{i}^{j(t)} = d\alpha_{i}^{j(q)} + (1-d)\alpha_{i}^{j(t)} \\
		&\widehat{\beta}_{i}^{j(t)} = d\beta_{i}^{j(q)} + (1-d)\beta_{i}^{j(t)},
	\end{aligned}
	\right. 
\end{equation}
where $q$ denotes the beginning epoch we set to reserve the fine prior information for the auxiliary model  $g(\bm{x}; \bm{\Omega})$, and $d\in (0,1)$ is also a positive constant used to provide the present model information $\alpha_{i}^{j(t)}, \beta_{i}^{j(t)}$ with the prior information $\alpha_{i}^{j(q)}, \beta_{i}^{j(q)}$. Before the epoch $r$, $\widehat{\lambda}_{i}^{j(t)}=\lambda_{i}^{j(t)}$ if $j\in S_i$, otherwise $1+\epsilon$. Before the epoch $q$, $\widehat{\alpha}_{i}^{j(t)}=\alpha_{i}^{j(t)}$ and $\widehat{\beta}_{i}^{j(t)}=\beta_{i}^{j(t)}$. In the above way, we refine the prior information epoch by epoch. 

The optimization of the two model $f(\bm{x}; \bm{\Theta})$ and $g(\bm{x}; \bm{\Omega})$ at the epoch $t$ for an instance $\bm{x}$ is as follows. First, $f(\bm{x}; \bm{\Theta})$ outputs $\bm{\lambda}$ while $g(\bm{x}; \bm{\Omega})$ outputs $\bm{\alpha}$ and $\bm{\beta}$, based on which, according to Eq.(\ref{theta_c}) and Eq.(\ref{z_c}), we replace $\bm{\theta}$ and $\bm{z}$ with the estimation $\widehat{\bm{\theta}}$ and $\widehat{\bm{z}}$ in Eq.(\ref{L_map_final}). Next, to introduce prior information, we use $\widehat{\bm{\lambda}}$, $\widehat{\bm{\alpha}}$ and $\widehat{\bm{\beta}}$ in Eq.(\ref{e_lambda}) and Eq.(\ref{e_alpha_beta}) to replace $\bm{\lambda}$, $\bm{\alpha}$ and $\bm{\beta}$ in Eq.(\ref{L_map_final}). Note that $\widehat{\bm{\theta}}$ and $\widehat{\bm{z}}$ are variables, of which we calculate the gradient to perform backpropagation, while $\widehat{\bm{\lambda}}$, $\widehat{\bm{\alpha}}$ and $\widehat{\bm{\beta}}$ are constants. Finally, we update the predictive model $f(\bm{x}; \bm{\Theta})$ and the auxiliary model $g(\bm{x}; \bm{\Omega})$ by fixing one and updating the other. The whole algorithmic description of the {\plmap} is shown in Algorithm \ref{alg:algorithm}. After implementing {\plmap} in the PLL training dataset, we can use the output $\bm{\lambda}$ of the main branch inference model $f(\bm{x};\bm{\Theta})$ to calculate the predict results $\widehat{\bm{\theta}}$ as the label confidence in the test dataset.

\begin{algorithm}[t]
	\caption{{\plmap} Algorithm}
	\label{alg:algorithm}
	\textbf{Input}: PLL training dataset $\mathcal{D} = \{(\bm{x}_{i}, S_{i}| 1 \leq i \leq n\}$, Epoch $\mathcal{T}$, Iteration $\mathcal{K}$; \\
	\textbf{Output}: The predictive model $f(\bm{x}; \bm{\Theta})$
	\begin{algorithmic}[1] 
		\STATE Initialize the parameters of the predictive model $f(\bm{x}; \bm{\Theta})$ and the auxiliary model $g(\bm{x}; \bm{\Omega})$;
		\STATE Initialize $\widehat{\bm{\lambda}}^{(1)}_i$,$\widehat{\bm{\alpha}}^{(1)}_i$,$\widehat{\bm{\beta}}^{(1)}_i$ for each instance $\bm{x}_i$;
		\FOR{$t=1,2,\dots,\mathcal{T}$}
		\STATE Randomly shuffle the training dataset $\mathcal{D}$ and divide it into $\mathcal{K}$ mini-batches;
		\FOR{$k=1,2,\dots,\mathcal{K}$}
		
		\STATE Calculate the parameters of Categorical distribution and Bernoulli distribution $\widehat{\bm{\theta}}_{i}$ and $\widehat{\bm{z}}_{i}$ for the instance $\bm{x}_i$ by the Eq.(\ref{theta_c}) and (\ref{z_c});
		\STATE Calculate the parameters $\widehat{\bm{\lambda}}^{(t)}_i$, $\widehat{\bm{\alpha}}^{(t)}_i$, $\widehat{\bm{\beta}}^{(t)}_i$ for the instance $\bm{x}_i$ by the Eq.(\ref{e_lambda}) and (\ref{e_alpha_beta});
		\STATE Fix $\bm{\Theta}$, update $\bm{\Omega}$ by the Eq.(\ref{L_map_final});
		\STATE Fix $\bm{\Omega}$, update $\bm{\Theta}$ by the Eq.(\ref{L_map_final});
		\ENDFOR
		\ENDFOR
	\end{algorithmic}
	
\end{algorithm}

\section{Theoretical Analysis}

In this section, we pay attention to the estimation error bound of the predictive model $f(\bm{x}; \bm{\Theta})$. According to Eq.(\ref{L_map_final}), the empirical risk estimator for the predictive model $f(\bm{x}; \bm{\Theta})$ is denoted by $\widehat{R}(f)=\frac{1}{n}\sum_{i=1}^{n}\mathcal{L}_{\text{MAP}}\left(f(\bm{x}_i),S_i\right)$ and for further analysis, we give an upper bound of $\mathcal{L}_{\text{MAP}}\left(f(\bm{x_i}),S_i\right)$ as follows:
\begin{equation}\label{upper_bound}
	\begin{aligned}
		\mathcal{L}_{\text{MAP}}\left(f(\bm{x}_i),S_i\right) \leq -\left(K_i + \sum_{j=1}^{c}w_{i}^{j}\ell(f(\bm{x}_i),\bm{e}^{j})\right),
	\end{aligned}
\end{equation}
where $\bm{e}^{j}$ denotes the standard canonical vector in $\mathbb{R}^{c}$, $\ell$ denotes the cross-entropy function, $w_{i}^j=\lambda_{i}^{j} -1 + \frac{1}{|S_i|}$ if $j\in S_i$, otherwise $w_{i}^{j}=\lambda_{i}^{j} -1$, and $K_i=\log |S_i| + \frac{1}{|S_i|}\sum_{j\in S_i}\log \prod_{k\in \overline{S}_i^j}z_i^k\prod_{k\notin \overline{S}_i^j}(1-z_i^k) +(\alpha_i^j-1)\log{z_i^j}+(\beta_i^j-1)\log{(1-z_i^j)}$. We denote $K = \max\{K_1,K_2,...,K_n\}$ and scale $w_{i}^j$ to $[0, \rho]$ during the training process. The detailed induction of Eq.(\ref{upper_bound}) can be seen in Appendix \ref{D_eq12}.
Then to formulate the estimation error bound of $f$, we give the following definition and lemmas.
\begin{definition}\label{definition}
	Let $\bm{x}_1, \bm{x}_2, \dots, \bm{x}_n \in \mathcal{X}$ be $n$ i.i.d random variables drawn from a probability distribution and $\sigma_1, \sigma_2,\dots,\sigma_n\in\{-1,+1\}$ be Rademacher variables with even probabilities, $\mathcal{H}=\{h:\mathcal{X} \mapsto \mathbb{R}\}$ be a class of measurable functions. Then the expected Rademacher complexity of $\mathcal{H}$ is defined as 
	$$
	\mathfrak{R}_n(\mathcal{H}) = \mathbb{E}_{\bm{x}, \sigma}\left[\sup_{h\in\mathcal{H}}\frac{1}{n}\sum_{i=1}^{n}\sigma_i h(\bm{x}_i)\right] .
	$$
\end{definition}

According Definition \ref{definition}, given the function space $\mathcal{G}=\left\{(\bm{x},S)\mapsto \mathcal{L}_{\text{MAP}}\left(f(\bm{x}),S\right) |f\in\mathcal{F}\right\}$, the expected Rademacher complexity of $\mathcal{G}$ can be defined as follows:
\begin{equation}
	\widetilde{\mathfrak{R}}_n(\mathcal{G}) = \mathbb{E}_{\bm{x},S, \sigma}\left[\sup_{g\in\mathcal{G}}\frac{1}{n}\sum_{i=1}^{n}\sigma_i g(\bm{x}_i, S_i)\right] .
\end{equation} 

We pre-limit both the main model $f$ and the auxiliary model $g$ by clamping their output to $[-A, A]$, and the loss $\mathcal{L}_{\text{MAP}}$ can be bounded, though it would not have extended to infinity in practice.

\begin{lemma}\label{lemma1}
	Suppose the loss function $\mathcal{L}_{\text{MAP}}\left(f(\bm{x}),S\right)$ is bounded by $M$,i.e.,$M=\sup_{\bm{x}\in\mathcal{X},S \in \mathcal{C},f\in\mathcal{F}}\mathcal{L}_{\text{MAP}}\left(f(\bm{x}),S\right)$, then for any $\xi\, \textgreater\, 0$, with probability at least $1-\xi$,
	$$
	\sup_{f\in\mathcal{F}}\left| R(f)-\widehat{R}(f)\right| \leq 2\mathfrak{R}_n(\mathcal{G})+\frac{M}{2}\sqrt{\frac{\log \frac{2}{\xi}}{2n}}.
	$$
\end{lemma}

\begin{lemma}\label{lemma2}
	Assume the loss function $\ell\left(f(\bm{x}), \bm{e}^{\iota}\right)$ is $L$-Lipschitz with respect to $f(\bm{x})\,(0\,\textless\, L\, \textless\, \infty)$ for all $\iota \in \mathcal{Y}$. Let $\mathcal{H}_{\iota}=\{h:\bm{x}\mapsto f_{\iota}(\bm{x}) \lvert f\in \mathcal{F}\}$ and $\mathfrak{R}_n(\mathcal{H}_{\iota}) = \mathbb{E}_{\bm{x}, \sigma}\left[\sup_{h\in\mathcal{H}_{\iota}}\frac{1}{n}\sum_{i=1}^{n}h(\bm{x}_i)\right]$, then the following inequality holds:
	$$
	\widetilde{\mathfrak{R}}_n(\mathcal{G})\leq \sqrt{2}\rho cL\sum_{\iota\in\mathcal{Y}}\mathfrak{R}_n(\mathcal{H}_{\iota}) + K,
	$$
\end{lemma}
The proof of Lemma \ref{lemma1} and \ref{lemma2} is provided in Appendix \ref{P_lemma1} and \ref{P_lemma2}. Based on Lemma \ref{lemma1} and \ref{lemma2}, we induce an estimation error bound for our {\plmap} method. Let $\widehat{f} = \arg\min_{f\in\mathcal{F}}\widehat{R}(f)$ be the empirical risk minimizer and $f^{\star}=\arg\min_{f\in\mathcal{F}}R(f)$ be the true minimizer. The function space $\mathcal{H}_{\iota}$ for the label $\iota\in \mathcal{Y}$ is defined as $\{h:x \mapsto f_{y}(x)|f\in\mathcal{F}\}$. Let $\mathfrak{R}_n(\mathcal{H}_{\iota})$ be the expected Rademacher complexity of $\mathcal{H}_{\iota}$ with sample size $n$, then we have the following theorem.

\begin{theorem}\label{theorem}
	Assume the loss function $\ell(f(\boldsymbol{x}), \bm{e}^\iota)$ is $L$-Lipschitz with respect to $f(\boldsymbol{x})\,(0\, \textless\, L\, \textless\, \infty)$ for all $\iota \in \mathcal{Y}$ and $\mathcal{L}_{\text{MAP}}\left(f(\bm{x}),S\right)$ is bounded by $M$, i.e.,$M=\sup_{\bm{x}\in\mathcal{X}, S\in\mathcal{C},f\in\mathcal{F}}\mathcal{L}_{\text{MAP}}\left(f(\bm{x}),S\right)$,. Then, for any
	$\xi>0$, with probability at least $1-\xi$,
	$$
	\begin{aligned}
		R(\widehat{f})-R \left(f^{\star}\right) \leq 4 \sqrt{2} &\rho cL \sum_{\iota\in \mathcal{Y}} \mathfrak{R}_{n}\left(\mathcal{H}_{\iota}\right)
		+M \sqrt{\frac{\log \frac{2}{\xi}}{2 n}} + 4K .
	\end{aligned}
	$$
\end{theorem}
The proof is provided in Appendix \ref{P_theorem1}. Theorem \ref{theorem} means that when $n \rightarrow \infty$ the empirical risk minimizer $f$ will converge to the optimal risk minimizer $f^{\star}$ as to all parametric models with a bounded norm.

\section{Experiments}

In this section, we validate the effectiveness of our proposed {\plmap} by performing it on benchmark datasets and real-world datasets and comparing its results against DNN-based PLL algorithms. Furthermore, the ablation study and sensitive analysis of parameters are conducted to explore {\plmap}.

\begin{table*}[t]
	\centering
	\small
	\setlength\tabcolsep{4pt}
	\caption{ Classification accuracy (mean$\pm$std) of each comparing approach on  benchmark datasets for instance-dependent PLL }
	\vspace{-0.5em}
	\begin{tabular}{cccccc}
		\toprule
        Datasets & MNIST &   FMNIST &  KMNIST &  CIFAR10 & CIFAR100 \\ 
        \midrule
        {\plmap} & $\bm{98.92\pm0.05\%}$ & $\bm{91.48\pm0.32\%}$ & $\bm{96.88\pm0.17\%}$ & $\bm{87.57\pm0.26\%}$ & $\bm{64.59\pm0.17\%}$ \\ 
        \midrule
        {\plcr} & 98.56$\pm$0.08\%$\bullet$ & 90.10$\pm$0.21\%$\bullet$ & 95.29$\pm$0.21\%$\bullet$ & 86.37$\pm$0.38\%$\bullet$ & 64.12$\pm$0.23\%$\bullet$ \\ 
        {\pico} & 98.61$\pm$0.12\%$\bullet$ & 88.41$\pm$0.20\%$\bullet$ & 94.78$\pm$0.19\%$\bullet$ & 86.16$\pm$0.21\%$\bullet$ & 62.98$\pm$0.38\%$\bullet$ \\ 
        {\vpll} & 98.72$\pm$0.05\%$\bullet$ & 90.63$\pm$0.30\%$\bullet$ & 96.19$\pm$0.75\% & 85.48$\pm$0.62\%$\bullet$ & 62.96$\pm$0.96\%$\bullet$ \\ 
        {\cavl} & 98.84$\pm$0.05\%$\bullet$ & 87.94$\pm$0.19\%$\bullet$ & 93.69$\pm$0.28\%$\bullet$ & 59.67$\pm$3.30\%$\bullet$ & 52.59$\pm$1.01\%$\bullet$ \\ 
        {\lws} &  98.56$\pm$0.06\%$\bullet$ & 88.99$\pm$0.26\%$\bullet$ & 92.27$\pm$1.03\%$\bullet$ & 37.49$\pm$2.82\%$\bullet$ & 53.98$\pm$0.99\%$\bullet$ \\ 
        {\rcpll} & 98.41$\pm$0.09\%$\bullet$ & 89.60$\pm$0.19\%$\bullet$ & 93.78$\pm$0.17\%$\bullet$ & 85.95$\pm$0.40\%$\bullet$ & 63.41$\pm$0.56\%$\bullet$ \\ 
        {\ccpll} & 98.16$\pm$0.14\%$\bullet$ & 89.86$\pm$0.11\%$\bullet$ & 94.08$\pm$0.35\%$\bullet$ & 79.96$\pm$0.99\%$\bullet$ & 62.40$\pm$0.84\%$\bullet$ \\ 
        {\proden} & 98.39$\pm$0.10\%$\bullet$ & 89.79$\pm$0.24\%$\bullet$ & 93.79$\pm$0.24\%$\bullet$ & 86.04$\pm$0.21\%$\bullet$ & 62.56$\pm$1.49\%$\bullet$ \\ 
		\bottomrule
	\end{tabular}
	\label{benchmark_uniform}
\end{table*}

\begin{table*}[t]
	\centering
	\small
	\caption{Classification accuracy (mean$\pm$std) of comparing algorithms on  the real-world datasets.  }
	\vspace{-0.5em}
	\begin{tabular}{cccccc}
		\toprule
		&  Lost    & BirdSong     & MSRCv2            & Soccer Player     &     Yahoo!News \\
		\midrule
		{\plmap}     & \textbf{77.02$\pm$0.82}\%  & \textbf{74.23$\pm$0.17\%} & \textbf{50.45$\pm$0.47\%}& 55.99$\pm$0.28\% & \textbf{66.62$\pm$0.19\%} \\
		\midrule
		{\vpll}     & 76.87$\pm$0.86\%  & 73.39$\pm$0.26\%$\bullet$ & 49.97$\pm$0.43\%$\bullet$& 55.81$\pm$0.10\% & 66.26$\pm$0.13\%$\bullet$ \\
		{\cavl}     & 75.89$\pm$0.42\%$\bullet$  & 73.47$\pm$0.13\%$\bullet$ & 44.73$\pm$0.96\%$\bullet$& 54.06$\pm$0.67\%$\bullet$ & 65.44$\pm$0.23\%$\bullet$ \\
		{\lws}      & 73.13$\pm$0.32\%$\bullet$ & 51.45$\pm$0.26\%$\bullet$ & 49.85$\pm$0.49\%$\bullet$ & 50.24$\pm$0.45\%$\bullet$ & 48.21$\pm$0.29\%$\bullet$ \\
		{\rcpll}    & 76.26$\pm$0.46\%  & 69.33$\pm$0.32\%$\bullet$ & 49.47$\pm$0.43\%$\bullet$ & \textbf{56.02$\pm$0.59\%} & 63.51$\pm$0.20\%$\bullet$ \\
		{\ccpll}    & 63.54$\pm$0.25\%$\bullet$  & 69.90$\pm$0.58\%$\bullet$ & 41.50$\pm$0.44\%$\bullet$ & 49.07$\pm$0.36\%$\bullet$ & 54.86$\pm$0.48\%$\bullet$\\
		{\proden}   & 76.47$\pm$0.25\%  & 73.44$\pm$0.12\%$\bullet$ & 45.10$\pm$0.16\%$\bullet$  & 54.05$\pm$0.15\%$\bullet$ & 66.14$\pm$0.10\%$\bullet$ \\
		\bottomrule
	\end{tabular}
	\vspace{-1.5em}
	\label{real-world datasets}

\end{table*}
%

\subsection{Datasets}

We implement {\plmap} with compared DNN-based algorithms on five widely used benchmark datasets in deep learning, including \texttt{MNIST} \cite{lecun1998gradient}, \texttt{Kuzushiji-MNIST} \cite{clanuwat2018deep}, \texttt{Fashion-MINIST} \cite{xiao2017fashion}, \texttt{CIFAR-10} and \texttt{CIFAR-100} \cite{krizhevsky2009learning}. Instance-dependent partial labels for these datasets are generated through the same strategy as \cite{xu2021instance}, which for the first time consider instance-dependent PLL.


Besides, the comparing algorithms are also performed on five frequently used real-world datasets, which come from different practical application domains, including \texttt{Lost} \cite{cour2011learning}, \texttt{BirdSong} \cite{briggs2012rank}, \texttt{MSRCv2}  \cite{liu2012conditional}, \texttt{Soccer Player} \cite{zeng2013learning} and \texttt{Yahoo!News} \cite{guillaumin2010multiple}. 

For benchmark datasets, we split $10\%$ samples from the training datasets for validating. For each real-world dataset, we run the methods with $80\%$/$10\%$/$10\%$ train/validation/test split. Then we run five trials on each datasets with different random seeds and report the mean accuracy and standard deviation of all comparing algorithms.

\subsection{Baselines}

We compare {\plmap} with eight DNN-based methods:1) {\plcr} \cite{wu2022revisiting}, a regularized training framework which is based on data augmentation and utilizes the manifold consistency regularization term to preserve the manifold structure both in feature space and label space.  2) {\pico} \cite{wang2022pico}, a contrastive learning framework which is based on data augmentation and performs label disambiguation based on the contrastive prototypes.  3) {\vpll} \cite{xu2021instance}, an instance-dependent PLL framework which guides the training process via the recovered latent label distributions. 4) {\cavl} \cite{zhang2021exploiting}, a discriminative approach which identifies correct labels from candidate labels by class activation value. 5) {\lws} \cite{wen2021leveraged}, an identification-based method which introduces a leverage parameter to consider the trade-off between losses on candidate and non-candidate labels. 6) {\rcpll}  \cite{feng2020provably}, a risk-consistent PLL method which is induced by an importance reweighting strategy. 7) {\ccpll} \cite{feng2020provably}, a classifier-consistent PLL method which leverages the transition matrix describing the probability of the candidate label set given a correct label. 8) {\proden} \cite{lv2020progressive}, a self-training style algorithm which provides a framework to equip arbitrary stochastic optimizers and models in PLL. Note that {\plcr} and {\pico} will not be compared on the real-world datasets due to the requirement of data augmentation. 

To ensure that the comparisons are as fair as possible, we employ the same network backbone, optimizer and data augmentation strategy for all the comparing methods. For \texttt{MNIST}, \texttt{Kuzushiji-MNIST} and \texttt{Fashion-MNIST}, we take LeNet-5 as their backbone. For \texttt{CIFAR-10} and \texttt{CIFAR-100}, the network backbone is changed to ResNet-32 \cite{he2016deep}. For all the real-world datasets, we simply adopt the linear model. The optimizer is stochastic gradient descent (SGD) \cite{robbins1951stochastic} with momentum $0.9$ and batch size $256$. The details of data augmentation strategy are shown in Appendix \ref{data_aug}. Besides, the learning rate is selected from $\{10^{-4}, 10^{-3}, 10^{-2}\}$ and the weight decay are selected from $\{10^{-5}, 10^{-4}, 10^{-3}, 10^{-2}\}$ according to the performance on the validation.

\subsection{Experimental Results}

The performance of each DNN-based method on each corrupted benchmark dataset is summarized in Table \ref{benchmark_uniform}, where the best results are highlighted in bold and $\bullet$/$\circ$ indicates whether {\plmap} statistically wins/loses to the comparing method on each dataset additionally (pairwise t-test at 0.05 significance level). We can overall see that {\plmap} significantly outperforms all comparing approaches on all benchmark datasets (except on \texttt{Kuzushiji-MNIST} where {\vpll} performs comparably against {\plmap}), and the improvements are particularly noticeable on \texttt{Fashion-MNIST} and \texttt{CIFAR-10}.

Table \ref{real-world datasets} demonstrates the ability of {\plmap} to solve the PLL problem in real-world datasets. {\plcr} and {\pico} are not compared on the real-world PLL datasets due to the inability of data augmentation to be employed on the extracted features from various domains. We can find that our method has stronger competence than others in all datasets except \texttt{Soccer Player} where {\plmap} loses to {\rcpll} but still ranks second. As for \texttt{BirdSong}, \texttt{MSRCv2} and \texttt{Yahoo!News}, the performance of {\plmap} is significantly better than all other comparing algorithms. And when it comes to \texttt{Lost}, our method is comparable to {\vpll}, {\proden} and {\rcpll}, while obviously better than the rest. 

\begin{table}[t]
	\centering
	\small
	\caption{Classification accuracy (mean$\pm$std) for comparison against {\plml}.}
	\vspace{-0.5em}
	\label{Ablation}
	\begin{tabular}{lcc}
		\hline
		\hline
		Dataset         & {\plmap} & {\plml} \\
		\hline
		Lost            & \textbf{77.02$\pm$0.82\%} & 52.50$\pm$3.20\%$\bullet$ \\
		BirdSong        & \textbf{74.23$\pm$0.17\%} & 71.13$\pm$0.45\%$\bullet$ \\
		MSRCv2          & \textbf{50.45$\pm$0.47\%} & 43.04$\pm$0.98\%$\bullet$ \\
		Soccer Player   & \textbf{55.99$\pm$0.28\%} & 49.56$\pm$0.31\%$\bullet$ \\
		Yahoo!News      & \textbf{66.62$\pm$0.19\%} & 51.48$\pm$0.07\%$\bullet$ \\
		\hline
		\hline
	\end{tabular}
    \vspace{-1.5em}
\end{table}
\begin{figure*}[t]
	\centering
	\small
	\subfigure[\texttt{Lost}]{
		\includegraphics[scale=0.4]{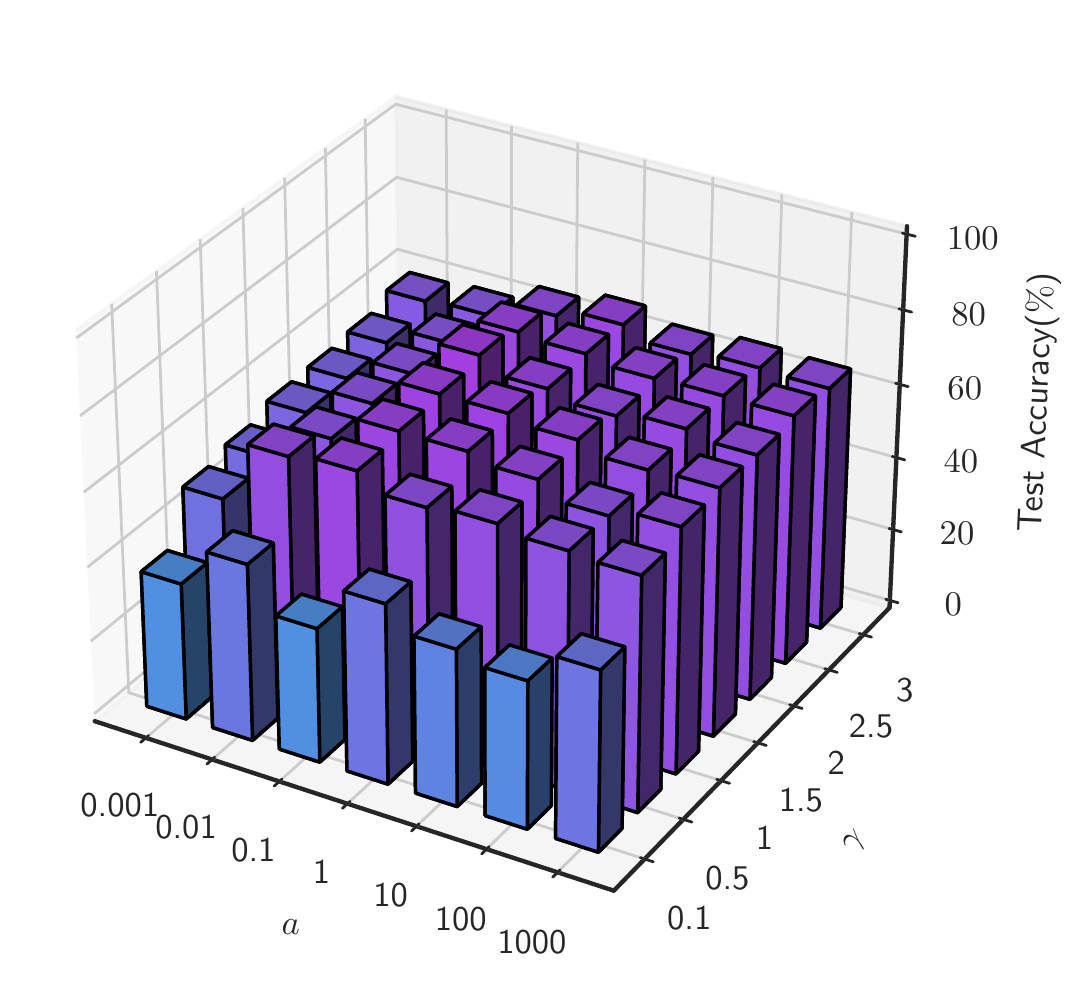}\label{4}
	}
	\hfill
	\subfigure[\texttt{BirdSong}]{
		\includegraphics[scale=0.4]{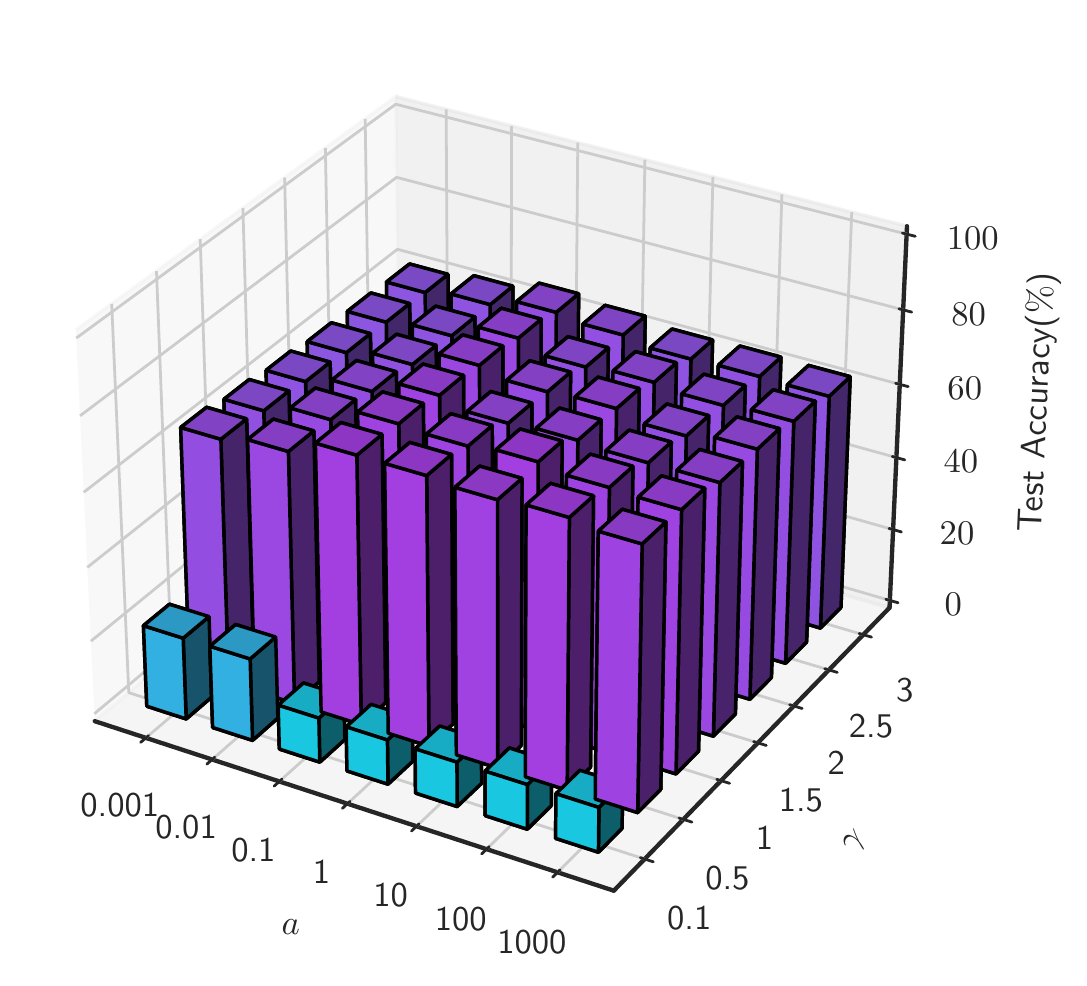}\label{3}
	}
	\hfill
	\subfigure[\texttt{Yahoo!News}]{
		\includegraphics[scale=0.4]{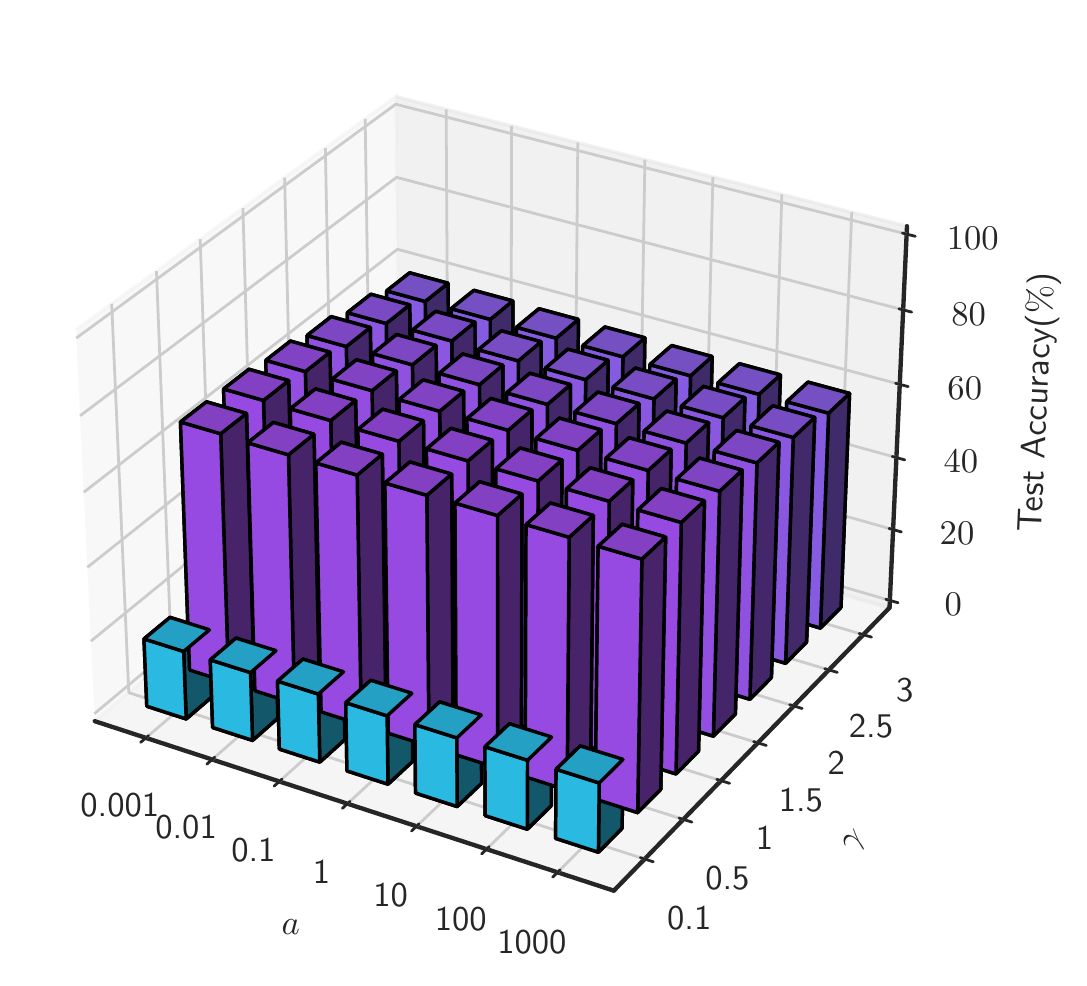}\label{4}
	}
	\hfill
	\vspace{-1em}
	\caption{Parameter sensitivity analysis for {\plmap} on \texttt{Lost}, \texttt{BirdSong} and \texttt{Yahoo!New}}
	\label{sensitive_analysis}
	\vspace{-1.5em}
\end{figure*}

\subsection{Further Analysis}

To demonstrate the effectiveness of the iteratively refined prior information introduced by {\plmap}, we remove the loss function $\mathcal{L}_{\text{reg}}$ to reverse the {\plmap} to {\plml} which only uses the log-likelihood function for optimization. The performance of {\plml} against {\plmap} is also measured by the classification accuracy (with pairwise t-test at 0.05 significance level). As is illustrated in Table \ref{Ablation}, {\plmap} achieves superior performance on all real-world datasets compared to {\plml} with the assistance of the prior information which {\plmap} provides and improves epoch by epoch.

Furthermore, we conduct parameter sensitivity analysis to study the influence of the two hyper-parameters $a$, $\gamma$ on our algorithm, which decides the scale of Dirichlet and Beta distribution parameters. Figure \ref{sensitive_analysis} illustrates the sensitivity of {\plmap} in the real-world datasets including \texttt{Lost}, \texttt{BirdSong}, \texttt{Yahoo!New} when $a$ varies from $0.001$ to $1000$ and $\gamma$ increases from $0.1$ to $3$. We can easily find that as for the small-scale real-world datasets like \texttt{Lost}, $a$ and $\gamma$ are suggested around $0.1$ and $0.5$, respectively. For the more large-scale real-world datasets like \texttt{BirdSong}, \texttt{Yahoo!News},  {\plmap} seems insensitive to $a$, and $\gamma$ are suggested around $1$.

\section{Conclusion}

In this paper, we consider a more realistic scenario, instance-dependent PLL, and explicitly decompose and model the generation process of instance-dependent candidate labels. Then based on the decompositional generation process, a novel instance-dependent PLL approach {\plmap} is proposed by us to further introduce and refine the prior information in every training epoch via MAP. The experimental comparisons with other DNN-based algorithms on both instance-dependent corrupted benchmark datasets and real-world datasets demonstrate the effectiveness of our proposed method.

\subsubsection*{Acknowledgments}

This research was supported by the National Key Research \& Development Plan of China (No. 2018AAA0100104), the National Science Foundation of China (62206050, 62125602, and 62076063), China Postdoctoral Science Foundation (2021M700023), Jiangsu Province Science Foundation for Youths (BK20210220), Young Elite Scientists Sponsorship Program of Jiangsu Association for Science and Technology (TJ-2022-078).

\bibliography{iclr2023_conference}
\bibliographystyle{iclr2023_conference}

\appendix
\section{Appendix}
\subsection{Derivations of Eq.(\ref{L_ml}) and Eq.(\ref{L_reg_w})}\label{D_eq4_6}
The derivation of Eq.(\ref{L_ml}) is as follows:
\begin{equation}
    \begin{aligned}
    \mathcal{L}_{\text{ML}} &=-\sum_{i=1}^{n} \log p(S_{i}| \bm{x}_i, \bm{\theta}_i, \bm{z}_i) \\
    &=-\sum_{i=1}^{n} \log \sum_{j\in S_i} p(y_i = j | \bm{x}_i)p(\bar{S}_i^j | \bm{x}_i) \\
    &=-\sum_{i=1}^{n} \log \sum_{j \in S_i} (Cat(\bm{l}_i | \bm{x}_i , \bm{\theta}_i) \prod_{k=1}^{c}Ber(\bar{s_i^k|z_i^k}) | y_i=j) \\
    &=-\sum_{i=1}^{n} \log \sum_{j \in S_i} ( \prod_{k=1}^{c} (\theta_i^k)^{l_i^k} \prod_{k=1}^{c} (z_i^k)^{\bar{s_i^k}}(1-z_i^k)^{1-\bar{s_i^k}}| y_i=j) \\
    &=-\sum_{i=1}^{n} \log \sum_{j \in S_i} \theta_i^j \prod_{k\in \bar{S_i^j}} z_i^k \prod_{k \notin \bar{S_i^j} } (1 - z_i^k).
    \end{aligned}
\end{equation}
The derivation of Eq.(\ref{L_reg_w}) is as follows:
\begin{equation}
    \begin{aligned}
    \mathcal{L}_{\text{reg}} &= - \sum_{i=1}^{n} \log p(\bm{\theta}_i | \bm{x}_i) + \log p(\bm{z}_i | \bm{x}_i) \\
    &=- \sum_{i=1}^{n} \log \frac{\Gamma(\sum_{j=1}^{c}\lambda_{i}^j)}{\prod_{j=1}^{c}\Gamma(\lambda_i^j)}\prod_{j=1}^{c} (\theta_i^j)^{\lambda_i^j-1} + \log \prod_{j=1}^{c} \frac{\Gamma (\alpha_i^j + \beta_i^j)}{ \Gamma (\alpha_i^j) \Gamma (\beta_i^j)} (z_i^j)^{\alpha_i^j -1} (1-z_i^j)^{\beta_i^j - 1} \\
    &=- \sum_{i=1}^{n} \log \prod_{j=1}^{c}(\theta_i^j)^{\lambda_i^j -1} + \log\prod_{j=1}^{c} (z_i^j)^{\alpha_i^j - 1} (1-z_i^j)^{\beta_i^j - 1} + C_{\Gamma} \\
    &=- \sum_{i=1}^{n} \sum_{j=1}^{c} (\lambda_i^j -1) \log (\theta_i^j)  + (\alpha_i^j -1) \log (z_i^j) + (\beta_i^j - 1)  \log (1 - z_i^j) + C_{\Gamma}. \\
    \end{aligned}
\end{equation}

$C_{\Gamma}=- \sum_{i=1}^{n} \log \frac{\Gamma(\sum_{j=1}^{c}\lambda_{i}^j)}{\prod_{j=1}^{c}\Gamma(\lambda_i^j)} +  \log \prod_{j=1}^{c} \frac{\Gamma (\alpha_i^j + \beta_i^j)}{ \Gamma (\alpha_i^j) \Gamma (\beta_i^j)}$. Due to that we later use $\hat{\bm{\lambda}}$, $\hat{\bm{\alpha}}$ and $\hat{\bm{\beta}}$ in Eq.(10) and Eq.(11) to replace $\bm{\lambda}$, $\bm{\alpha}$ and $\bm{\beta}$ in Eq.(7), which are fixed to be constants, $C_{\Gamma}$ will be a constant and ignored. Here, the derivation of Eq.(6) has been finished.

\subsection{Degeneration of Eq.(\ref{L_map_final})}\label{D_eq7}
The proposed process can accommodate the uniform generation process of candidate labels by settng the parameters of Bernoulli distribution to a constant $p$, which also means the flipping probability. In this case, we can degenerate $\mathcal{L}_{\text{ML}}, \mathcal{L}_{\text{reg}}$ and $\mathcal{L}$ to the following form.
\begin{equation}
    \begin{aligned}
    \mathcal{L}_{\text{ML-D}} &= -\sum_{i=1}^{n} {\log} \sum_{j \in S_i} \theta_i^j \prod_{k\in \bar{S_i^j}} z_i^k \prod_{k \notin \bar{S_i^j} } (1 - z_i^k) \\
    &= -\sum_{i=1}^{n} {\log} \sum_{j \in S_i} \theta_i^j  (1-p)^{c+1-|S_i|} p^{|S|-1}  \\
    &= -\sum_{i=1}^{n} {\log} (1-p)^{c+1-|S_i|} p^{|S|-1}  \sum_{j \in S_i} \theta_i^j  \\
    &= -\sum_{i=1}^{n} {\log}  \sum_{j \in S_i} \theta_i^j - \sum_{i=1}^{n}  {\log} (1-p)^{c+1-|S_i|} p^{|S|-1}.
    \end{aligned}
\end{equation}

Due to that the second term is a constant, the final $\mathcal{L}_{\text{ML}}$ is formulated as 

\begin{equation}
    \begin{aligned}
    \mathcal{L}_{\text{ML-D}} = -\sum_{i=1}^{n} {\log}  \sum_{j \in S_i} \theta_i^j.
    \end{aligned}
\end{equation}

Due to the parameters of Bernoulli Distribution are constant, we do not need Beta Distribution. Hence,
\begin{equation}
    \begin{aligned}
    \mathcal{L}_{\text{reg-D}} = -\sum_{i=1}^{n} \sum_{j=1}^{c} (\lambda_i^j -1 ) {\log} \theta^i_j.
    \end{aligned}
\end{equation}

Finally, 
\begin{equation}
    \begin{aligned}
    \mathcal{L}_{\text{D}} = \mathcal{L}_{\text{ML-D}} + \mathcal{L}_{\text{reg-D}} = -\sum_{i=1}^{n} {\log}  \sum_{j \in S_i} \theta_i^j -\sum_{i=1}^{n} \sum_{j=1}^{c} (\lambda_i^j -1 ) {\log} \theta^i_j.
    \end{aligned}
\end{equation}

\subsection{Derivations of Eq.(8) and Eq.(9)}\label{D_eq8_9}
\begin{equation}
    \begin{aligned}
        \bm{\theta}_i \sim p(\bm{\theta}_i | \bm{x}_i, \bm{\lambda}_i) &\propto Dir(\bm{\theta}_i | \bm{\lambda}_i) \cdot Cat(\bm{l}_i | \bm{\theta}_i) \\
        &=\frac{\Gamma(\sum_{j=1}^{c} \lambda_{i}^{j})}{\prod_{j=1}^{c}\Gamma(\lambda_{i}^{j})}\cdot \prod_{j=1}^{c} (\theta_{i}^{j})^{\lambda_{i}^{j} - 1} \cdot \prod_{j=1}^{c}(\theta_{i}^{j})^{l_{i}^{j}} \\
        &=\frac{\Gamma(\sum_{j=1}^{c} \lambda_{i}^{j})}{\prod_{j=1}^{c}\Gamma(\lambda_{i}^{j})}\cdot \prod_{j=1}^{c} (\theta_{i}^{j})^{\lambda_{i}^{j} + l_{i}^{j} - 1} \\
        &=Dir(\bm{\theta}_i | \bm{\lambda}_i + \bm{l}_i).
    \end{aligned}
\end{equation}
The above has proved the conjugacy of Dirichlet and Categorical distribution. For Dirichlet distribution , its expectation can be calculated as
\begin{equation}
    \begin{aligned}
        \mathbb{E}[\theta_i^j | \bm{x}_i, \bm{\lambda}_i] = \frac{\lambda_i^j + l_i^j}{\sum_{k=1}^{c} \lambda_{i}^{k} + l_i^k} = \frac{\lambda_i^j + o_i^j}{\sum_{k=1}^{c} \lambda_{i}^{k} + o_i^k}.
    \end{aligned}
\end{equation}
The mathematical derivations of Eq.(8) are completed. Eq.(9) can be proved in a similar way due to Dirichlet distribution and Categorical distribution are the generalization forms of Beta distribution and Bernoulli distribution respectively.

\subsection{Calculation Details of Eq. (12)}\label{D_eq12}
According to multivariate basic inequality, we can obtain:
\begin{equation}
	\begin{aligned}
		-\log\sum_{j\in S_i}\theta_i^j\prod_{k\in \overline{S}_i^j}z_i^k\prod_{k\notin \overline{S}_i^j}(1-z_i^k) \leq - \log |S_i| - \frac{1}{|S_i|}\sum_{j\in S_i}\left(\log\theta_i^j + \log \prod_{k\in \overline{S}_i^j}z_i^k\prod_{k\notin \overline{S}_i^j}(1-z_i^k)\right).
	\end{aligned}
\end{equation}
Then we can calculate Eq. (12) as
\begin{equation}
	\begin{aligned}
		&\mathcal{L}_\text{{MAP}}\left(f(\bm{x}_i),S_i\right) = -\log\sum_{j\in S_i}\theta_i^j\prod_{k\in \overline{S}_i^j}z_i^k\prod_{k\notin \overline{S}_i^j}(1-z_i^k) \\
		&- \sum_{j=1}^{c}(\lambda_i^j-1)\log{\theta_i^j} +(\alpha_i^j-1)\log{z_i^j}+(\beta_i^j-1)\log{(1-z_i^j)} \\
		\leq&- \log |S_i| - \frac{1}{|S_i|}\sum_{j\in S_i}\left(\log\theta_i^j + \log \prod_{k\in \overline{S}_i^j}z_i^k\prod_{k\notin \overline{S}_i^j}(1-z_i^k)\right)\\
		&-\sum_{j=1}^{c}(\lambda_{i}^{j}-1)\log\theta_{i}^{j}+(\alpha_i^j-1)\log{z_i^j}+(\beta_i^j-1)\log{(1-z_i^j)}\\
		=& -\left(\log |S_i|+\frac{1}{|S_i|}\sum_{j\in S_i}\log \prod_{k\in \overline{S}_i^j}z_i^k\prod_{k\notin \overline{S}_i^j}(1-z_i^k) + \sum_{j=1}^{c} (\alpha_i^j-1)\log{z_i^j}+(\beta_i^j-1)\log{(1-z_i^j)} \right)  \\
		&-\sum_{j\in S_i}\left(\lambda_{i}^{j}-1+\frac{1}{|S_i|}\right)\log\theta_i^j - \sum_{j\in S_i}\left(\lambda_{i}^{j}-1\right)\log\theta_i^j \\
		=& -\left(K_i + \sum_{j=1}^{c}w_{i}^{j}\ell(f(\bm{x}_i),\bm{e}^{j})\right)
	\end{aligned}
\end{equation}
where $\bm{e}^{j}$ denotes the standard canonical vector in $\mathbb{R}^{c}$, and $\ell$ denotes the cross-entropy function, $w_{i}^j=\lambda_{i}^{j} -1 + \frac{1}{|S_i|}$ if $j\in S_i$, otherwise $w_{i}^{j}=\lambda_{i}^{j} -1$, and $K_i=\log |S_i| + \frac{1}{|S_i|}\sum_{j\in S_i}\log \prod_{k\in \overline{S}_i^j}z_i^k\prod_{k\notin \overline{S}_i^j}(1-z_i^k) +(\alpha_i^j-1)\log{z_i^j}+(\beta_i^j-1)\log{(1-z_i^j)}$.

\subsection{Proof of Lemma 1}\label{P_lemma1}

In order to prove this lemma, we first show $\mathcal{L}_{\text{MAP}}$ can be bounded. Due to the ouput of $f, g$ is limited in $[-A, A]$, the following inequations hold:
\begin{equation}
\widehat{\theta}_i^j \geq \frac{a\exp(-A / {\gamma}) + b}{ac\exp(A / \gamma ) +bc+c}=B
\end{equation} 

\begin{equation}
E=  \frac{a\exp(-A / {\gamma}) + b}{2(a\exp(A / {\gamma}) + b) + 1}  \leq \widehat{z}_i^j \leq  \frac{a\exp(A / {\gamma}) + b+1}{a\exp(A / {\gamma})+a\exp(-A / {\gamma})+2b + 1} = F
\end{equation} 

Hence, 
$$
\mathcal{L}_{\text{MAP}} \leq - \log |S_i| B(1-F)^{c+1-|S_i|}E^{|S_i|-1} - c \log [BE(1-F)]^{a\exp(A/{\gamma}) + b}
$$

When $M$ takes the value larger than $- {\log} |S_i| B(1-F)^{c+1-|S_i|}E^{|S_i|-1} - c {\log} [BE(1-F)]^{a\exp(A/{\gamma}) + b}$, the loss $\mathcal{L}_{\text{MAP}}$ will be bounded. Note that the limitation to the output of the model excludes extreme (conditional) probabilities, the effect of which could be ignored which $A$ is large enough.

Then, we show that the one direction $\sup_{f\in\mathcal{F}}R(f)-\widehat{R}(f)$ is bounded with probability at least $1 - {\xi}/{2}$, and the other direction can be similarly shown. Suppose an example $(\bm{x}_i, S_i)$ is replaced by another arbitrary example  $(\bm{x}'_i, S'_i)$, then the change of $\sup_{f\in\mathcal{F}}R(f)-\widehat{R}(f)$ is no greater than ${M}/{(2n)}$, the loss function $\mathcal{L}$ are bounded by $M$. By applying McDiarmid's inequality, for any $\xi \textgreater 0$, with probability at least $1 - \xi/2$,
\begin{equation}
	\sup_{f\in\mathcal{F}}R(f)-\widehat{R}(f) \leq \mathbb{E}\left[\sup_{f\in\mathcal{F}} R(f)-\widehat{R}(f) \right] + \frac{M}{2} \sqrt{\frac{\log\frac{2}{\delta}}{2n}}.
\end{equation} 

By sysmetrization, we can obtain
\begin{equation}
	\mathbb{E}\left[\sup_{f\in\mathcal{F}} R(f)-\widehat{R}(f) \right] \leq 2\widetilde{\mathfrak{R}}_n(\mathcal{G}).
\end{equation}
By further taking into account the other side $\sup_{f\in\mathcal{F}} R(f)-\widehat{R}(f)$, we have for any $\xi \textgreater 0$, with probability at least $1 - \xi$, 
\begin{equation}
	\sup_{f\in\mathcal{F}}\lvert R(f)-\widehat{R}(f)\rvert \leq 2\mathfrak{R}_n(\mathcal{G})+\frac{M}{2}\sqrt{\frac{\log \frac{2}{\xi}}{2n}}
\end{equation}

\subsection{Proof of Lemma 2}\label{P_lemma2}
The upper bound loss function of $\mathcal{L}_{\text{{MAP}}}(f(\bm{x_i}), S_i)$ is denoted by
\begin{equation}
	\bar{\mathcal{L}}(f(\bm{x_i}), S_i) = -\left(K_i + \sum_{j=1}^{c}w_{i}^{j}\ell(f(\bm{x}_i),\bm{e}^{j})\right)
\end{equation}
Correspondingly, the function space for $\bar{\mathcal{L}}$ can be defined as:
\begin{equation}
	\bar{\mathcal{G}}=\left\{(\bm{x},S)\mapsto \bar{\mathcal{L}}\left(f(\bm{x}),S\right) |f\in\mathcal{F}\right\}
\end{equation}
Then the expected Rademacher complexity of $\bar{\mathcal{G}}$ can be defined as follows:
\begin{equation}
	\widetilde{\mathfrak{R}}_n(\bar{\mathcal{G}}) = \mathbb{E}_{\bm{x},S, \sigma}\left[\sup_{\bar{g}\in\bar{\mathcal{G}}}\frac{1}{n}\sum_{i=1}^{n}\sigma_i \bar{g}(\bm{x}_i, S_i)\right]
\end{equation} 

For each example $(\bm{x}_i, S_i)$, since $w_i^j$ is bounded in $[0, \rho]$ and $K_i\leq K$, we can obtain $\widetilde{\mathfrak{R}}_n(\mathcal{G}) \leq \widetilde{\mathfrak{R}}_n(\bar{\mathcal{G}}) \leq \rho c\mathfrak{R}_n(\ell \circ \mathcal{F}) + K$ where $\ell \circ \mathcal{F}$ denotes $\{\ell \circ \mathcal{F} | f\in \mathcal{F}\}$. Since $\mathcal{H}_{\iota} = \{ h: \bm{x} \mapsto f_{y}(\bm{x})|f\in \mathcal{F}\}$ and the loss function $\ell\left(f(\bm{x},\bm{e}^{\iota})\right)$ is $L$-Lipschitz for all $\iota \in \mathcal{Y}$, by the Rademacher vector contraction inequality, we have $\mathfrak{R}_n(\ell \circ \mathcal{F})\leq \sqrt{2}L\sum_{\iota \in \mathcal{Y}}\mathfrak{R}_n(\mathcal{H}_{\iota})$. Then the proof is completed.

\subsection{Proof of Theorem 1}\label{P_theorem1}
Theorem is proven through
\begin{equation}
	\begin{aligned}
		R(\widehat{f}) - R(f^{*}) &= R(\widehat{f}) - \widehat{R}(\widehat{f}) + \widehat{R}(\widehat{f}) - \widehat{R}(f^{*}) + \widehat{R}(f^{*}) - R(f^{*}) \\
		&\leq R(\widehat{f}) - \widehat{R}(\widehat{f}) + \widehat{R}(f^{*}) - R(f^{*}) \\ 
		&\leq 2\sup_{f\in\mathcal{F}}\left|{R(f)-\widehat{R}(f)}\right| \\
		&\leq 4\widetilde{\mathfrak{R}}_n(\mathcal{G}) + M \sqrt{\frac{\log\frac{2}{\delta}}{2n}} \\
		&\leq 4\sqrt{2}\rho cL\sum_{\iota \in \mathcal{Y}}\mathfrak{R}_n(\mathcal{H}_{\iota}) + M \sqrt{\frac{\log\frac{2}{\delta}}{2n}} + 4K
	\end{aligned}
\end{equation}

\subsection{Details of data augmentation}\label{data_aug}

For all benchmark datasets, we apply Random Horizontal Flipping, Random Cropping, and Cutout. For 
\texttt{CIFAR-10} and \texttt{CIFAR-100}, AutoAugment is additionally applied.

\subsection{Extending Experiments}\label{extra_experiment}
\textbf{Our Generation Model}. We also synthesize the sampled correct label from Categorical Distribution $p( y=j | \bm{x})$ and incorrect candidate labels from Bernoulli Distribution $p(\bar{s}^j=1 | \bm{x})$. For the former, correct labels has been contained in these datasets. For the latter, we use the confidence prediction of a clean neural network $g(\bm{x}; \hat{\bm{\Omega}})$ (trained only with correct labels) to model the Bernoulli Distribution, i.e., $p(\bar{s}^j=1 | \bm{x})=\text{sigmoid}(g_j(\bm{x}; \hat{\bm{\Omega}}))$. Table \ref{benchmark_ours} illustrates the performance of {\plmap} and comparing approaches on on benchmark datasets, of which instance-dependent partial labels generated by our generation model. From the table, we can overall see that {\plmap} consistently outperforms all comparing approaches on all benchmark datasets, and the improvements are particularly noticeable on \texttt{Kuzushiji-MNIST} and \texttt{CIFAR-10}.

\begin{table*}[t]
	\centering
	\small
	\setlength\tabcolsep{4pt}
	\caption{ Classification accuracy (mean$\pm$std) of each comparing approach on benchmark datasets, of which instance-dependent partial labels are generated by our generation model.}
	\vspace{-0.5em}
	\begin{tabular}{cccccc}
		\toprule
        Datasets & MNIST &   FMNIST &  KMNIST &  CIFAR10 & CIFAR100 \\ 
        \midrule
        {\plmap} & $\bm{98.87\pm0.05\%}$ & $\bm{91.20\pm0.21\%}$ & $\bm{97.28\pm0.19\%}$ & $\bm{90.18\pm0.32\%}$ & $\bm{64.67\pm0.23\%}$ \\ 
        \midrule
        {\plcr} & 98.46$\pm$0.22\%$\bullet$ & 90.18$\pm$0.27\%$\bullet$ & 95.19$\pm$0.13\%$\bullet$ & 87.07$\pm$0.34\%$\bullet$ & 64.43$\pm$0.44\% \\ 
        {\pico} & 98.61$\pm$0.12\%$\bullet$ & 88.32$\pm$0.17\%$\bullet$ & 94.78$\pm$0.19\%$\bullet$ & 86.56$\pm$2.74\%$\bullet$ & 63.05$\pm$0.26\%$\bullet$ \\ 
        {\vpll} & 98.77$\pm$0.04\%$\bullet$ & 90.58$\pm$0.36\%$\bullet$ & 96.41$\pm$0.33\%$\bullet$ & 86.31$\pm$0.37\%$\bullet$ & 63.77$\pm$0.51\%$\bullet$ \\ 
        {\cavl} & 98.77$\pm$0.09\%$\bullet$ & 88.09$\pm$0.32\%$\bullet$ & 93.30$\pm$0.55\%$\bullet$ & 60.20$\pm$2.64\%$\bullet$ & 52.85$\pm$2.21\%$\bullet$ \\ 
        {\lws} &  98.52$\pm$0.14\%$\bullet$ & 88.90$\pm$0.25\%$\bullet$ & 92.69$\pm$0.76\%$\bullet$ & 39.75$\pm$2.22\%$\bullet$ & 53.52$\pm$1.37\%$\bullet$ \\ 
        {\rcpll} & 98.35$\pm$0.14\%$\bullet$ & 89.78$\pm$0.29\%$\bullet$ & 94.06$\pm$0.05\%$\bullet$ & 85.48$\pm$0.27\%$\bullet$ & 63.24$\pm$0.97\%$\bullet$ \\ 
        {\ccpll} & 98.15$\pm$0.07\%$\bullet$ & 89.67$\pm$0.31\%$\bullet$ & 94.12$\pm$0.17\%$\bullet$ & 80.19$\pm$0.76\%$\bullet$ & 62.11$\pm$1.03\%$\bullet$ \\ 
        {\proden} & 98.43$\pm$0.10\%$\bullet$ & 89.72$\pm$0.36\%$\bullet$ & 94.08$\pm$0.45\%$\bullet$ & 85.60$\pm$0.20\%$\bullet$ & 62.87$\pm$0.45\%$\bullet$ \\ 
		\bottomrule
	\end{tabular}
	\label{benchmark_ours}

\end{table*}

\begin{table}[t]
	\centering
	\small
	\caption{Classification accuracy (mean$\pm$std) for comparison against {\plml} on benchmark datasets, of which instance-dependent partial labels are generated by our generation model.}
	\vspace{-0.5em}
	\label{Ablation_2}
	\begin{tabular}{lcc}
		\hline
		\hline
		Dataset         & {\plmap} & {\plml} \\
		\hline
		MNIST           & \textbf{98.87$\pm$0.05\%} & 91.79$\pm$0.12\%$\bullet$ \\
		Kuzushiji-MNIST & \textbf{97.28$\pm$0.19\%} & 76.67$\pm$0.34\%$\bullet$ \\
		Fashion-MINIST  & \textbf{91.27$\pm$0.21\%} & 78.38$\pm$0.29\%$\bullet$ \\
		CIFAR-10        & \textbf{90.18$\pm$0.32\%} & 69.47$\pm$0.45\%$\bullet$ \\
		CIFAR-100       & \textbf{64.67$\pm$0.23\%} & 40.23$\pm$0.57\%$\bullet$ \\
		\hline
		\hline
	\end{tabular}
    \vspace{-1.5em}
\end{table}

\begin{figure*}[b]
	\centering
	\small
	\subfigure[\texttt{Fashion-MINIST}]{
		\includegraphics[scale=0.4]{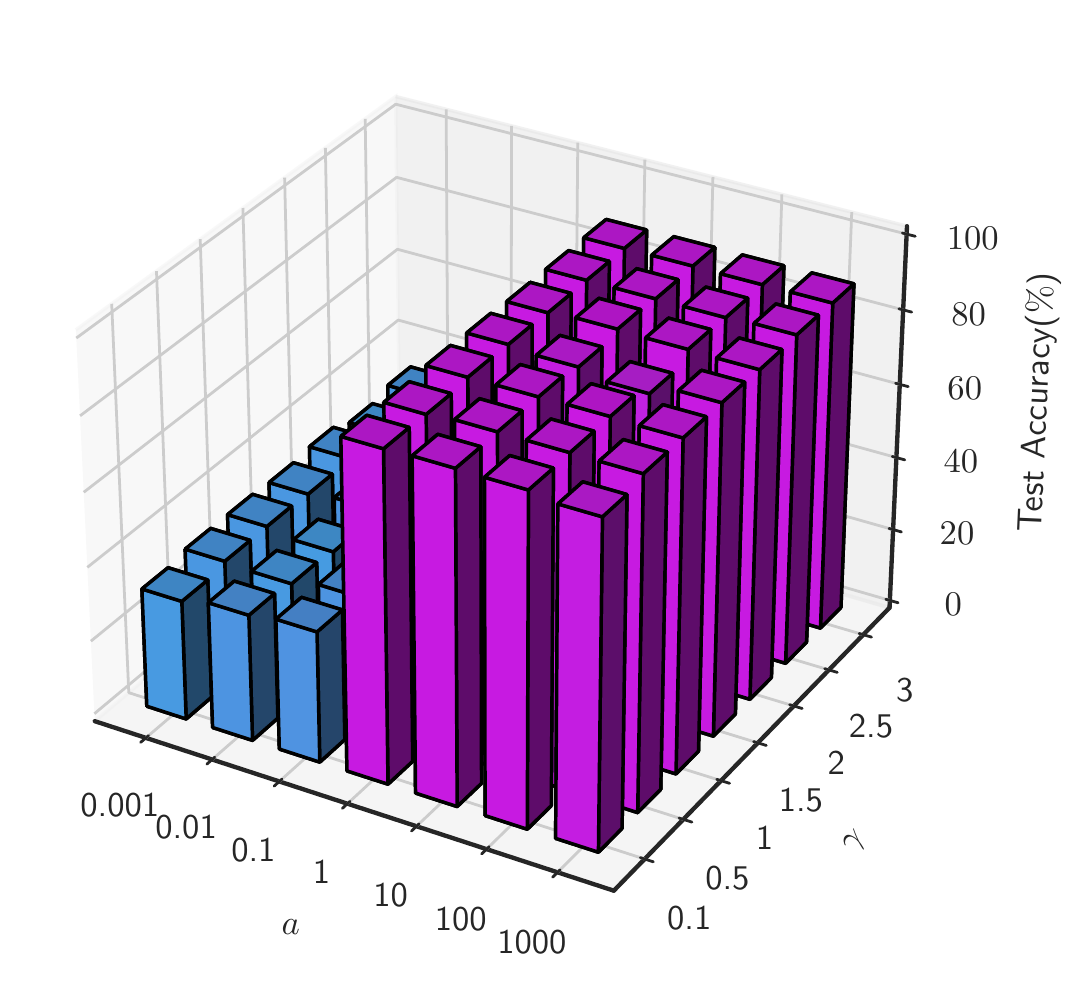}\label{4}
	}
	\vspace{-1em}
	\caption{Parameter sensitivity analysis for {\plmap} on \texttt{Fashion-MINIST}.}
	\label{sensitive_analysis}
	\vspace{-1.5em}
\end{figure*}

\textbf{Ablation Study}. As is illustrated in Table \ref{Ablation_2}, {\plmap} also achieves superior performance on all benchmark datasets, of which instance-dependent partial labels are generated by our generation model.

\textbf{Sensitive Analysis}. For the benchmark dataset \texttt{Fashion-MINIST}, the performance of {\plmap} is stable and effective in the case that $a$ is around $10$ and $\gamma$ is around $1.5$.

\end{document}